\documentclass[runningheads]{llncs}

\usepackage{eccv}
\usepackage{eccvabbrv}

\usepackage{graphicx}
\usepackage{booktabs}
\usepackage{multirow}
\usepackage{epsfig}
\usepackage{amsmath}
\usepackage{amssymb}
\usepackage{ulem}
\usepackage{xcolor}
\usepackage{stmaryrd}
\usepackage{hyperref}
\newif\ifshowchanges

\showchangesfalse

\ifshowchanges

  \newcommand{\remSylv}[1]{\textcolor{red}{\sout{#1}}}
  \newcommand{\addSylv}[1]{\textcolor{black}{#1}}
  \newcommand{\comSylv}[1]{\textcolor{orange}{\textit{#1}}}

  \newcommand{\comSte}[1]{\textcolor{purple}{\textit{#1}}}
  \newcommand{\addSte}[1]{\textcolor{black}{#1}}
  \newcommand{\remSte}[1]{\textcolor{gray}{\sout{#1}}}

  \newcommand{\comVic}[1]{\textcolor{teal}{\textit{#1}}}
  
  \newcommand{\remVic}[1]{\textcolor{teal}{\sout{#1}}}

  \newcommand{\addEmi}[1]{\textcolor{black}{#1}}

\else

  \newcommand{\remSylv}[1]{}
  \newcommand{\addSylv}[1]{\textcolor{black}{#1}}
  \newcommand{\comSylv}[1]{}

  \newcommand{\comSte}[1]{}
  \newcommand{\addSte}[1]{\textcolor{black}{#1}}
  \newcommand{\remSte}[1]{}

  \newcommand{\comVic}[1]{}
  
  \newcommand{\remVic}[1]{}

  \newcommand{\addEmi}[1]{#1}

\fi

\newcommand{\orcidlink}[1]{}
\begin{document}

\title{ 
Uncertainty-Aware Trajectory Forecasting from Imperfect Tracking}

 \titlerunning{Uncertainty-Aware Trajectory Forecasting}

\author{Stephane Da Silva Martins\inst{1}\orcidlink{0000-1111-2222-3333} \and
Victor Petrovic\inst{2}\orcidlink{1111-2222-3333-4444} \and
Emanuel Aldea\inst{1}\orcidlink{1111-2222-3333-4444} \and
Sylvie Le Hégarat-Mascle\inst{1}\orcidlink{2222--3333-4444-5555}}

\authorrunning{S. Da Silva Martins et al.}

\institute{
 SATIE -- CNRS UMR 8029, Université Paris-Saclay, France
 \email{\{stephane.da-silva-martins, emanuel.aldea,
sylvie.le-hegarat\}@universite-paris-saclay.fr}
 \and
 ENS Paris-Saclay, Université Paris-Saclay, France
 \email{victor.petrovic@ens-paris-saclay.fr}
}

\maketitle

\begin{abstract}

\addSte{
\addSylv{Most t}rajectory forecasting models are 
trained 
\addSylv{on} clean annotated histories, \addSylv{and are often evaluated under the same idealized assumption,} although practical deployments 
\addSylv{rely on} trajectories produced by imperfect multi-object trackers. 
The 
\addSylv{real-world} observations exhibit localization jitter, missed or unstable detections, and data-association ambiguity, which are usually either ignored or removed 
\addSylv{through} denoising. This paper instead treats tracking-derived reliability cues as an informative signal to be propagated to the predictor. We propose a plug-in} uncertainty-aware 
\addSte{formulation 
\addSylv{in which} each observed state 
\addSylv{is encoded as an uncertain state representation, modeled by a Gaussian distribution} whose covariance combines detection-level localization uncertainty and association-level ambiguity through the law of total variance. 
\addSylv{Existing backbones are adapted with minimal architectural changes: input trajectories are represented as Gaussian observations, and predicted trajectories are produced as Gaussian forecasts rather than deterministic coordinates. }
To train predictors that remain robust under structured observation noise, we combine temporally correlated Ornstein-Uhlenbeck perturbations with} response-based knowledge distillation 
\addSte{from} a teacher trained on 
\addSte{clean trajectories. }
\addSte{Experiments on Oxford Town Centre and VIRAT using real tracker outputs, together with a complementary ETH/UCY pseudo-detection protocol, show that the proposed formulation improves displacement accuracy and \addSte{the reliability–sharpness trade-off of probabilistic forecasts}
.}

 \end{abstract}

\section{Introduction}
\label{sec:intro}

Multi-agent trajectory 
\addSte{forecasting is a key component of robotic navigation, video understanding, and safety-critical decision-making systems. Modern predictors have made substantial progress on standard pedestrian benchmarks by modeling social interactions, scene context, and the multimodality of future motion using graph neural networks, Transformers, recurrent architectures, and generative models~\cite{SocialVAE,yuan2021agentformer,mart}
. Yet most of these methods 
\addSylv{implicitly assume} that the observed past trajectory is clean, temporally consistent, and directly available at inference time.}

\addSte{This assumption is rarely satisfied in 
\addSylv{operational settings}}. 
\addSte{
\addSylv{Observed histories are usually obtained from} an upstream multi-object tracker, whose outputs 
\addSylv{may contain} localization noise, confidence fluctuations, missed detections, and identity switches. 
\addSylv{Importantly, t}hese imperfections are 
\addSylv{not only sources of error; they also carry information about the reliability of the observations}: 
unstable 
\addSylv{detections, low-confidence boxes} or ambiguous association\addSylv{s} 
\addSylv{often reflect challenging} scene 
\addSylv{conditions} such as occlusion 
\addSylv{or} crowding. 
\addSylv{Removing or ignoring these signals may therefore discard useful information for the forecasting model}. 
Recent work on prediction from raw videos and robust trajectory forecasting has shown that tracking errors can 
\addSylv{strongly affect} downstream 
performance~\cite{yu2021towards,li2025natra}, motivating predictors that 
\addSylv{propagate} observation uncertainty 
\addSylv{to the predictor instead of treating the past trajectory as clean and deterministic}.}
\addSte{In this work, we specifically focus on localization uncertainty and association ambiguity at observed time steps. Explicit handling of missing observations and recovery from completed identity switches are outside the scope of the present formulation.}

\addSte{We address this problem by propagating tracking uncertainty through the forecasting pipeline. Instead of representing the past as deterministic points, each observation is modeled as a Gaussian state whose mean is the estimated pedestrian position and whose covariance \addSylv{matrix} 
\addSylv{quantifies the uncertainty about the pedestrian’s true position}. We decompose this covariance into two interpretable components: }\textit{\addSte{localization uncertainty}}\addSte{, induced by uncertain detections, and }\textit{\addSte{association uncertainty}}\addSte{, induced by multiple plausible matching candidates. These two terms are combined through the law of total variance, yielding a compact 
\addSylv{uncertainty} 
signal that can be consumed by existing trajectory forecasting backbones.}

\addSte{The resulting formulation is intentionally simple. A standard coordinate embedding is replaced by an uncertainty-aware 
embedding 
\addSylv{of the Gaussian state parameters} $(\mu_x,\mu_y,\sigma_x,\sigma_y,\rho_{xy})$, where $\rho_{xy}$ 
\addSte{encodes} the correlation 
\addSylv{term of the} covariance matrix, and the output head predicts Gaussian 
\addSylv{moments for the predicted positions,} optimized with a negative log-likelihood objective. This keeps the core predictor unchanged while allowing} the model  
\addSte{
\addSylv{to account for observations with large estimated uncertainty} and 
\addSylv{produce probabilistic forecasts}. 
During training, we simulate temporally correlated tracking errors with an Ornstein--Uhlenbeck process parameterized from empirical tracker residuals. We further use response-based knowledge distillation: a teacher trained on clean trajectories provides privileged 
\addSylv{predictions}, while the student learns from corrupted probabilistic observations. This encourages the student to 
\addSte{learn} motion dynamics \addSylv{closer to those inferred from clean trajectories,} without forcing it to copy the teacher's covariance. }

\noindent Our main contributions are summarized as follows:
\begin{enumerate}
\item
\addSte{We formulate trajectory forecasting 
\addSylv{using imperfect tracker outputs} as an uncertainty propagation problem, representing 
\addSylv{observed and predicted states} as Gaussian states rather than deterministic points}.
 \item We 
 \addSte{derive a tracker-output uncertainty estimate that separates localization uncertainty from association ambiguity and combines 
 \addSylv{these two components} through the law of total variance.
 }
 \item
 \addSte{We introduce an uncertainty-aware training strategy that couples temporally correlated noise injection with} response-based 
 \addSte{distillation from \addSylv{a teacher trained on} clean trajectories.}
 \item
 \addSte{We evaluate robustness and probabilistic reliability on real tracking outputs from Oxford Town Centre and VIRAT, and on a complementary ETH/UCY pseudo-detection protocol.
} \end{enumerate}

\section{Related Work}
\addSte{\subsection{Trajectory Prediction under Clean Observations}
Trajectory prediction has been extensively studied under the assumption that past trajectories are accurately observed. Prior work has explored several complementary aspects of this setting, including the modeling of social interactions between agents, the incorporation of scene context, and the extension of prediction horizons from short-term extrapolation to longer-range reasoning~\cite{alahi2016social,gupta2018social,kothari2022trajectory,zhang2022social,vista,mohamed2020social,yuan2021agentformer,cheng2023gatraj,xu2023eqmotion,na2023spu,bai2025sceneaware}. To address these challenges, the literature has progressively moved from recurrent and pooling-based models to graph-based, attention-based, and transformer-based architectures, and more recently to state-space models and semantic priors that improve long-range reasoning and scene awareness~\cite{ren2025stochastic,chib2025lg}.}

\addSte{Since human motion is inherently stochastic and multimodal, a large body of work has further modeled future uncertainty through generative mechanisms such as GANs~\cite{gupta2018social}, VAEs~\cite{SocialVAE,lee2022muse}, diffusion models~\cite{li2023bcdiff,gu2022stochastic}, and flow-matching methods~\cite{brinke2025flow}. These approaches are effective at representing multiple plausible futures conditioned on past motion, social interactions, or scene context. However, they generally assume that the observed past is clean and complete. Our work targets this complementary source of uncertainty: the reliability of the input trajectory itself, which may be noisy, partial, or corrupted in unconstrained perception settings.}

\subsection{Robust 
\addSte{Forecasting from Noisy Observations}}


\addSte{Robust trajectory forecasting 
\addSylv{addresses} the gap between benchmark annotations and real 
\addSylv{perception outputs}. One line of work bypasses or reduces the dependenc
\addSylv{e} on explicit tracking by predicting from raw videos, detections, or multi-hypothesis tracking structures~\cite{yu2021towards,zhang2023towards,weng2022mtp,weng2022whose}. Another line explicitly 
\addSylv{attempts to denoise} or disentangle 
corrupted histories before forecasting.} OosTraj~\cite{zhang2024oostraj} introduces a vision--positioning denoising module for out-of-sight trajectories, CaDeT~\cite{pourkeshavarz2024cadet} uses causal disentanglement, and NATRA~\cite{li2025natra}
\addSte{learns} noise-agnostic
\addSte{representations through mutual-information constraints. These methods 
\addSylv{share our} motivation \addSylv{of improving robustness to imperfect observations}, but they typically treat 
\addSylv{tracking errors} as a nuisance to remove. We instead expose the predictor to an uncertainty 
\addSylv{representation} derived from the tracking process, so that the model can 
\addSylv{modulate} its reliance on each observation \addSylv{according to its estimated uncertainty}}.

\subsection{\addSte{
\addSylv{Tracking Uncertainty and Privileged Distillation}}
}
\addSte{Uncertainty estimation is 
\addSylv{essential when trajectory forecasting relies on imperfect perception outputs}. 
In \addSylv{multi-object} tracking, localization covariance, detector confidence, and association ambiguity provide 
cues about \addSylv{the uncertainty of the estimated} state
~\cite{barshalom2001estimation,deepsort,botsort}. 
We use these cues to build a Gaussian observation 
\addSylv{representation by combining detection-level and association-level uncertainty} through moment matching. Knowledge distillation has also been used to transfer privileged or teacher information to a student model~\cite{hinton2015distilling,vapnik2009lupi,lopezpaz2016unifying}. 
Our setting follows this learning-using-privileged-information view. \addSylv{During training,} the teacher 
\addSylv{has access to} clean histories, while the student 
\addSylv{receives} corrupted \addSylv{probabilistic} histories\addSylv{. The distillation objective encourages the student to align its predictions with} 
clean 
\addSylv{history predictions, without forcing it to mimic the teacher's uncertainty estimate}.}

 \section{Methodology}
 \label{sec:method}

\addSte{Given a } pedestrian $i$, 
\addSte{a trajectory predictor observes a past sequence $\mathbf{X}_i=\{(x_t^i,y_t^i)\}_{t=1}^{T_{\text{obs}}}$ and forecasts future positions $\mathbf{Y}_i=\{(x_t^i,y_t^i)\}_{t=T_{\text{obs}}+1}^{T_{\text{obs}}+T_{\text{pred}}}$. 
\addSylv{In this paper}, we assume that $\mathbf{X}_i$ is produced by an upstream tracker and therefore comes with observation reliability that should be propagated to the forecasting model}.

\begin{figure}[t]
 \centering
 \includegraphics[width=\linewidth]{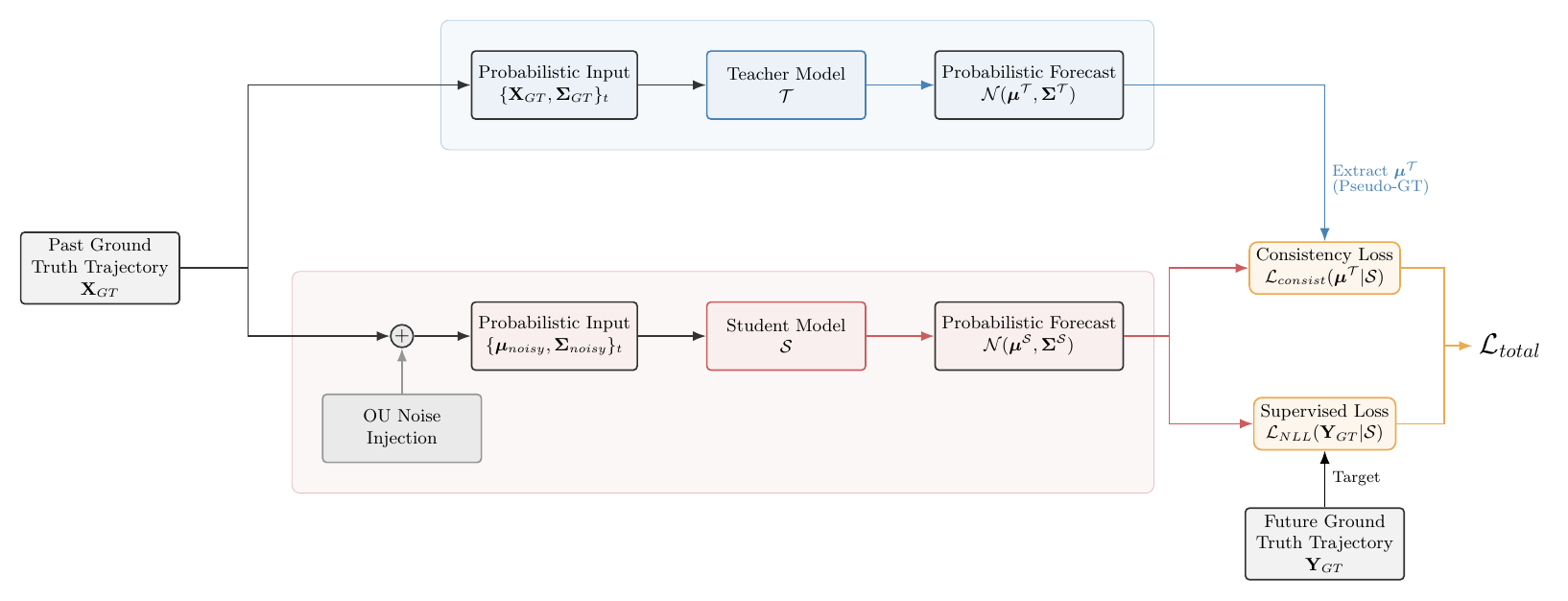}
 \caption{ 
\textbf{\addSte{Overview of} the \addSte{proposed uncertainty-aware training pipeline}.}
\addSte{A teacher model } is 
\addSte{trained on clean trajectories, whereas } a 
\addSte{student model receives } probabilistic 
\addSte{observations corrupted by temporally correlated tracking-like noise}. The 
\addSte{student } is 
optimized
\addSte{with } a supervised 
\addSte{negative log-likelihood } and a \addSte{response-based } consistency loss 
\addSte{on } the teacher 
\addSte{mean}.}
 \label{fig:pipeline}
 \end{figure}

\addSte{Our framework has three components. First, tracker outputs are converted into Gaussian observation 
\addSylv{representations} by combining detection-level and asso-ciation-level uncertainty. \addEmi{Second, we adapt existing backbones in order to 
\addSylv{process} these} 
\addSylv{representations} and to predict Gaussian 
\addSylv{parameters for future positions}. 
Third, training combines structured corruption and teacher--student distillation so that the student 
\addSylv{aligns its predictions made} 
from noisy observations \addSylv{with clean-history predictions,} while 
\addSylv{retaining its own} uncertainty 
\addSylv{estimates}
. Figure~\ref{fig:pipeline} summarizes this training pipeline.
}

\subsection{Probabilistic Modeling of Tracking Uncertainty}

Most predictors assume deterministic inputs $\mathbf{X}_i$. In real-world scenarios,
$\mathbf{X}_i$ is produced by an upstream MOT and
is corrupted by localization noise and association ambiguity.
We therefore
model the input as a sequence of random variables.

For clarity in the
following, we omit the pedestrian index $i$. Let
$\mathbf{x}_t$ $\in \mathbb{R}^2$
denote the random vector representing the true position of a pedestrian at time $t$.
Instead of representing the tracker output as a deterministic point, we model each observation
as a probability distribution
$p(\mathbf{x}_t)$
over positions (e.g., a bivariate Gaussian), parameterized from the tracker outputs.

\noindent\textbf{Law of Total Variance for Tracker Uncertainty.} At
time
$t$, the tracker
provides a set of $N$ candidate detections $\mathcal{D}_t = \{d_k\}_{k=1}^N$
. Each candidate detection $d_k$ is characterized by a 2D position $\boldsymbol{\mu}_k\in \mathbb{R}^2$, a bounding-box size $(W_k, H_k)$, and a confidence score $s_k$.

\noindent\textbf{Soft association over candidates.}
Let $c_k$ denote an association cost between the current track state and candidate $d_k$. From these costs, we define normalized association weights $(w_k)_{k=1}^N$ via a softmax with the temperature $\tau>0$:
\begin{equation}
\label{eq:softmax_weights}
w_k \triangleq \frac{\exp(-c_k/\tau)}{\sum_{j=1}^{N}\exp(-c_j/\tau)}, \qquad \text{so that} \quad
w_k \ge 0,\ \sum_{k=1}^N w_k = 1.
\end{equation}

In this study, we use $\tau=1$ in all experiments. Before applying the softmax, association costs are normalized per frame and per cost type, so that the fixed temperature is applied on comparable scales. We then introduce a discrete random variable $Z\in\llbracket 1,N\rrbracket$ representing the association hypothesis, with
$\mathbb{P}(Z=k) = w_k$.
Note that $(w_k)$ is a soft assignment induced by the costs (with $\tau$), not necessarily a calibrated posterior from the tracker.

\noindent\textbf{Observation model (detection noise).}
We model the observed 2D position at time $t$ as a random vector $\tilde{\mathbf{x}}_t \in \mathbb{R}^2$.
Conditioned on the association hypothesis $\{Z=k\}$, we assume
\begin{equation}
\label{eq:cond_gauss}
\tilde{\mathbf{x}}_t \mid \{Z=k\} \sim \mathcal{N}(\boldsymbol{\mu}_k, \boldsymbol{\Sigma}_k),
\end{equation}
where $\boldsymbol{\mu}_k$ is the detected position and $\boldsymbol{\Sigma}_k$ encodes the spatial uncertainty associated with detection candidate $d_k$.
Inspired by box-geometry cues used in standard trackers~\cite{deepsort} and recent attempts to modulate measurement noise with detection confidence~\cite{strongsort}, we use a confidence-aware localization proxy rather than a calibrated detector posterior. This deliberately simple mapping should not be interpreted as a calibrated estimate of detector uncertainty; it only provides a monotonic reliability cue, with larger boxes and lower-confidence detections producing broader spatial uncertainty. Detector-specific or calibrated covariance estimates could replace this proxy without changing the remainder of the framework. We set the covariance of each candidate as:
\begin{equation}
\label{eq:sigma_k}
\boldsymbol{\Sigma}_k =
\begin{pmatrix}
\frac{W_k^2}{s_k+\epsilon_s} & 0\\
0 & \frac{H_k^2}{s_k+\epsilon_s}
\end{pmatrix},
\end{equation}
so that low-confidence detections, for which $s_k$ is close to $0$, yield large uncertainty, whereas high-confidence detections, for which $s_k$ is close to $1$, keep the variance close to the object's squared spatial extent. Here, $W_k$ and $H_k$ denote the bounding-box dimensions and $s_k \in (0,1]$ denotes the detection confidence.

\noindent\textbf{Moment matching.}
The marginal distribution induced by Eqs.~\eqref{eq:softmax_weights} and~\eqref{eq:cond_gauss} is a Gaussian mixture, which we reduce to a single Gaussian by matching its first two moments. By total expectation,
\begin{equation}
\label{eq:mu_total}
\boldsymbol{\mu}_{\text{total}} = \mathbb{E}[\tilde{\mathbf{x}}_t]
= \sum_{k=1}^N w_k\,\boldsymbol{\mu}_k,
\end{equation}
and by total covariance,
\begin{equation}
\label{eq:total_variance}
\boldsymbol{\Sigma}_{\text{total}}
= \underbrace{\sum_{k=1}^N w_k\boldsymbol{\Sigma}_k}_{\boldsymbol{\Sigma}_{\text{pos}}}
+ \underbrace{\sum_{k=1}^N w_k(\boldsymbol{\mu}_k-\boldsymbol{\mu}_{\text{total}})(\boldsymbol{\mu}_k-\boldsymbol{\mu}_{\text{total}})^\top}_{\boldsymbol{\Sigma}_{\text{assoc}}}.
\end{equation}
Here, $\boldsymbol{\Sigma}_{\text{pos}}$ is the weighted localization uncertainty of the candidate detections, whereas $\boldsymbol{\Sigma}_{\text{assoc}}$ measures their spatial dispersion and therefore association ambiguity. In the single-candidate case, $\boldsymbol{\Sigma}_{\text{assoc}}=\mathbf{0}$; it also remains small for spatially concentrated candidates and increases when several separated candidates receive comparable weights. The two components are used to construct $\boldsymbol{\Sigma}_{\text{total}}$ and are not passed separately to the forecasting network.

\noindent Finally, we parameterize the input to the prediction model as
\begin{equation}
\tilde{\mathbf{x}}_t \approx \mathcal{N}(\boldsymbol{\mu}_{\text{total}}, \boldsymbol{\Sigma}_{\text{total}}).
\end{equation}

\noindent\textbf{\addSte{Unified uncertainty interface.}}
\addSte{Training-time perturbations and inference-time tracking ambiguity are expressed through the same covariance interface $\boldsymbol{\Sigma}_{\text{total}}$. The predictor therefore receives both the estimated position and an explicit descriptor of its reliability, without requiring the forecasting architecture to distinguish the underlying source of uncertainty.}

 \noindent\textbf{Black-box deployment: proxy association costs.}
In deployment, the Multi-Object Tracking (MOT) may be a black box and may not expose its internal association cost matrix. To retain a multi-hypothesis formulation, we reconstruct proxy costs $c'_k$ \textit{a posteriori} by comparing the current track state to each detection candidate $d_k\in\mathcal{D}_t$.
This formulation implicitly propagates uncertainty from $t-1$: as the track's variance or bounding box size increases, the resulting costs $c'_k$ (e.g., Mahalanobis or IoU) reflect a higher association ambiguity.
When the deployed tracker relies on specific metrics (e.g., Mahalanobis gating and ReID cosine distance as in DeepSORT~\cite{deepsort} or BoT-SORT~\cite{botsort}), we approximate these components to compute $c'_k$ and obtain weights $w_k$ via Eq.~\eqref{eq:softmax_weights}.

\noindent When appearance features are unavailable, we approximate association costs via two IoU-based proxies. \textit{Raw IoU} computes the overlap directly between the last observed track box and each detection candidate $d_k$. \textit{Projected IoU} first compensates for the track's motion by projecting its position forward using the estimated velocity at the last time step.

\subsection{ 
\addSte{Gaussian Observation Representation} and Input Embedding}

To incorporate the probabilistic input derived above into modern trajectory prediction backbones such as SingularTrajectory~\cite{singular}, VISTA~\cite{vista} or MART~\cite{mart}, we modify the input embedding layer. Standard models typically employ a linear projection layer $\phi: \mathbb{R}^2 \rightarrow \mathbb{R}^d$ to map deterministic 2D coordinates $\mathbf{x}_t = (x_t, y_t)$ into a $d$-dimensional latent feature space.

In our framework, the observation at time $t$ is parameterized by the Gaussian statistics $(\boldsymbol{\mu}_{\text{total}}, \boldsymbol{\Sigma}_{\text{total}})$. To obtain a compact uncertainty descriptor while preserving the 
\addSte{geometry } of $\boldsymbol{\Sigma}_{\text{total}}$, we 
\addSte{encode the covariance by its marginal standard deviations and correlation coefficient. Let}
\begin{equation}
\addSte{
\sigma_x = \sqrt{[\boldsymbol{\Sigma}_{\mathrm{total}}]_{xx}}, \qquad
\sigma_y = \sqrt{[\boldsymbol{\Sigma}_{\mathrm{total}}]_{yy}}, \qquad
\rho_{xy} = \frac{[\boldsymbol{\Sigma}_{\mathrm{total}}]_{xy}}{\sigma_x\sigma_y + \epsilon},
}
\end{equation}
\addSte{where $\epsilon$ is a small numerical constant and $\rho_{xy}$ is clipped to $[-1+\epsilon,1-\epsilon]$ when needed}. We then adapt the embedding layer to project 
\addSte{this five-dimensional Gaussian observation representation } into the same latent dimension $d$:

\begin{equation}
 \mathbf{E}_t = \phi_{\text{prob}}\left( \mu_x, \mu_y, \sigma_x, \sigma_y \addSte{, \rho_{xy} } \right) \in \mathbb{R}^d.
\end{equation}


By keeping the embedding dimension $d$ identical to that of the original backbone, our embedding $\mathbf{E}_t$ remains a plug-and-play modification. As a result, the proposed uncertainty-aware input can complement certain existing deterministic SOTA predictors with minimal architectural changes, without modifying their core forecasting modules. The predictor receives the combined statistics $(\boldsymbol{\mu}_{\text{total}},\boldsymbol{\Sigma}_{\text{total}})$; $\boldsymbol{\Sigma}_{\text{pos}}$ and $\boldsymbol{\Sigma}_{\text{assoc}}$ are not separate input channels.

\subsection{Probabilistic Output and Training Loss}
We preserve each backbone's native stochastic trajectory-generation mechanism and replace only its final two-dimensional position head by a five-dimensional Gaussian head. Consequently, every native future sample $k$ is represented at each prediction step $t$ by
$\mathcal{N}(\boldsymbol{\mu}_{\mathrm{pred},t}^{(k)},\boldsymbol{\Sigma}_{\mathrm{pred},t}^{(k)})$, where the mean $\boldsymbol{\mu}_{\mathrm{pred},t}^{(k)}=(\mu_x,\mu_y)$ replaces the original predicted position and $(\sigma_x,\sigma_y,\rho_{xy})$ parameterize its covariance. For SingularTrajectory, the diffusion sampling process is therefore unchanged; only the final mapping from each generated sample to a 2D position is replaced by this Gaussian head. The same principle is used for MART and VISTA. We optimize the Gaussian outputs with the negative log-likelihood (NLL):

\begin{equation}
 \mathcal{L}_{\text{NLL}} = \frac{1}{2} \sum_{t=1}^{T_{\text{pred}}} \left( \log |\boldsymbol{\Sigma}_{\text{pred},t}| + (\mathbf{y}_t - \boldsymbol{\mu}_{\text{pred},t})^\top (\boldsymbol{\Sigma}_{\text{pred},t})^{-1} (\mathbf{y}_t - \boldsymbol{\mu}_{\text{pred},t}) \right),
 \label{eq:nll_loss}
\end{equation}
where $\mathbf{y}_t$ denotes the ground-truth position. By minimizing $\mathcal{L}_{\text{NLL}}$, the model is trained not only to minimize displacement error but also to learn a covariance that captures predictive dispersion. Its empirical reliability is evaluated using coverage and area metrics.

\subsection{Robust Training via Empirical Noise Injection}

Training a model directly on outputs from a specific tracker
can imprint tracker-specific error patterns and limit generalization. Conversely, training only on clean data
does not expose the model to the
noise distribution it will face
at inference time. To bridge this gap, we propose a \textit{tracking-noise-aware training} strategy. Rather than relying on standard
i.i.d.
Gaussian noise, which ignores the temporal
structure of tracking errors, we simulate
realistic dynamics by coupling an empirical distribution with an \textit{Ornstein-Uhlenbeck} process.

\noindent\textbf{Empirical Error Distribution.}
To capture the stochastic properties of real-world tracking failures, we first analyze the error residuals of a baseline tracker evaluated on the training set. We apply kernel density estimation (KDE) to estimate the probability density function (PDF) of the
error magnitudes. During training, a base error scale $\sigma_i$ is sampled from this empirical distribution for each pedestrian $i$, so that the injected noise reflects different levels of tracking difficulty.

\noindent\textbf{Ornstein-Uhlenbeck Drift Process.}
Unlike i.i.d. Gaussian noise, real-world tracking errors often exhibit temporal correlation due to the inertia of filtering algorithms (e.g., Kalman filters). This
``colored noise'' effect
~\cite{barshalom2001estimation}
implies that an error at time $t$ is likely to
persist at $t+1$.

To replicate these dynamics, we model the error residuals $\mathbf{n}_t$ as an Ornstein-Uhlenbeck process
, defined by the following Stochastic Differential Equation (SDE):

\begin{equation}
 d\mathbf{n}_t = -\theta (\mathbf{n}_t - \mu\mathbf{1}_2) dt + \sigma_i\ d\mathbf{W}_t,
 \label{eq:ou_continuous}
\end{equation}
where:
\begin{itemize}
 \item $\theta$ is the mean-reversion rate (stiffness), pulling both spatial components of the error back towards the scalar mean $\mu$ (typically $\mu=0$ for an unbiased tracker
 ).
 \item $\sigma_i$ is the volatility (diffusion
 coefficient) specific to pedestrian $i$.
 \item $\mathbf{W}_t$ denotes a standard Wiener process (Brownian motion).
\end{itemize}

For numerical implementation during training, we apply the Euler-Maruyama discretization to Eq.~
\eqref{eq:ou_continuous}
with a time step $\Delta t$. The update rule for the noise injected into pedestrian $i$'s trajectory becomes:

\begin{equation}
 \mathbf{n}_{i,t+1} = \mathbf{n}_{i,t} - \underbrace{\theta (\mathbf{n}_{i,t}-\mu\mathbf{1}_2) \Delta t}_{\text{Drift}} + \underbrace{\sigma_i \sqrt{\Delta t} \cdot \boldsymbol{\xi}_{i,t}}_{\text{Diffusion}},
\end{equation}
where $\boldsymbol{\xi}_{i,t} \sim \mathcal{N}(\mathbf{0}, \mathbf{I}_2)$ is a standard Gaussian vector sampled at each step. This formulation generates smooth, temporally correlated error trajectories that mimic the drift behavior of imperfect trackers, consistent with recent trajectory simulation studies~\cite{langas2024human}.

\subsection{Privileged Knowledge Transfer via Response-Based Distillation}
\label{sec:dist}
To mitigate performance degradation
due to input corruption, we adopt a Knowledge Distillation (KD)~\cite{hinton2015distilling} strategy
within the Learning Using Privileged Information (LUPI) paradigm~\cite{vapnik2009lupi}. The
link between KD and LUPI~\cite{lopezpaz2016unifying}
has been used to bridge modality gaps and
improve model robustness
to noisy or missing inputs in vision tasks~\cite{hoffman2016learning, garcia2018modality}.
Here,
ground-truth trajectories are available during training
as privileged information, to guide the learning
of a model operating on imperfect
data at test time.

We use response-based rather than feature-based distillation~\cite{gou2021knowledge}. Matching only model outputs avoids architecture-specific mappings between intermediate representations and keeps the distillation mechanism portable across backbones.

\noindent{\textbf{Teacher-Student Formulation.}}
As illustrated in Figure~ \addSte{\ref{fig:pipeline}}, we instantiate a dual-network architecture comprising a \textit{Teacher} model $\mathcal{T}$ and a \textit{Student} model $\mathcal{S}$. While both share
the same backbone architecture
, they operate on
different inputs:
\begin{itemize}
 \item \textbf{Teacher $\mathcal{T}$:} Receives
 ground truth coordinates
 $\mathbf{X}_{\text{GT}}$. To
 keep the input interface consistent with the Student,
 we cast these inputs into a probabilistic form
: a Gaussian distribution centered on the ground truth with a small, fixed isotropic standard deviation (e.g., $\sigma = 0.1$ m) to
 reflect annotation noise. This constitutes the \textit{privileged information}
 and is available only during
 training.
 \item \textbf{Student $\mathcal{S}$:} Receives the probabilistic input $(\boldsymbol{\mu}_{\text{noisy}}, \boldsymbol{\Sigma}_{\text{noisy}})$ generated by the noise-injection module, simulating the uncertainty of real-world tracking outputs.
\end{itemize}

The Teacher is used only during training and is discarded at inference time. It is therefore a privileged-information training component rather than a competing deployable predictor, since its clean observation history is unavailable in the target setting. Since the Teacher $\mathcal{T}$ is trained on clean trajectories, it provides a predictive signal conditioned on clean observations. Distillation transfers this signal to the Student $\mathcal{S}$:
by minimizing a consistency loss that treats the Teacher's mean prediction as a pseudo-target, we
encourage the Student to
produce plausible futures that align with ground-truth dynamics, even if the
observations are corrupted.

\noindent\textbf{Objective Function and Predictive Uncertainty.}
Let $\mathcal{S}(\cdot \mid \mathbf{X}_{\text{noisy}})$ denote the Student model, which outputs a probabilistic forecast parameterized as a sequence of bivariate Gaussians $\mathcal{N}(\boldsymbol{\mu}_t^{\mathcal{S}}, \boldsymbol{\Sigma}_t^{\mathcal{S}})$ for each future time step $t$. Similarly, the Teacher $\mathcal{T}(\cdot \mid \mathbf{X}_{\text{GT}})$ predicts
Gaussians
$\mathcal{N}(\boldsymbol{\mu}_t^{\mathcal{T}}, \boldsymbol{\Sigma}_t^{\mathcal{T}})$.
In our distillation
training, we deliberately
use only the Teacher's
mean
prediction $\boldsymbol{\mu}_t^{\mathcal{T}}$ as a pseudo-target.

To transfer the Teacher's
clean predictive signal to the Student, the training objective combines
supervised learning
with a distillation constraint.
The total loss
$\mathcal{L}_{total}$ is
a weighted
sum:
\begin{equation}
 \mathcal{L}_{\text{total}} = \lambda_{\text{traj}}\,\mathcal{L}_{\text{NLL}}(\mathbf{Y}_{\text{GT}}\mid \mathcal{S}) + \lambda_{\text{dist}}\, \mathcal{L}_{\text{consist}}(\boldsymbol{\mu}^{\mathcal{T}} \mid \mathcal{S}),
\end{equation}
where
$\lambda_{\text{traj}}$ and $\lambda_{\text{dist}}$ balance the two objectives.

\noindent\textbf{Supervised Loss (
$\mathcal{L}_{\text{NLL}}$).} The first term is the
negative log-likelihood of the future ground truth $\mathbf{Y}_{\text{GT}}$ under the Student's predictive distribution, encouraging accurate forecasting and reliable aleatoric uncertainty.

\noindent\textbf{Consistency Distillation (
$\mathcal{L}_{\text{consist}}$).} The second term enforces consistency between the Student's probabilistic output and the Teacher's
mean
prediction.
We formulate
it as the (Gaussian) NLL of the Teacher's mean
under the Student's predictive distribution:

\begin{equation}
\label{eq:consistency_nll}
 \mathcal{L}_{\text{consist}}
 = \frac{1}{2} \sum_{t=1}^{T_{\text{pred}}}
 \left( \log \left|\boldsymbol{\Sigma}_t^{\mathcal{S}}\right|
 + (\boldsymbol{\mu}_t^{\mathcal{T}} - \boldsymbol{\mu}_t^{\mathcal{S}})^\top
 (\boldsymbol{\Sigma}_t^{\mathcal{S}})^{-1}
 (\boldsymbol{\mu}_t^{\mathcal{T}} - \boldsymbol{\mu}_t^{\mathcal{S}})
 \right).
\end{equation}
We use this NLL-based consistency
rather than a full Kullback-Leibler (KL) divergence between the Teacher's and Student's predictive distributions. A KL divergence would
encourage the Student to match the Teacher's covariance ($\boldsymbol{\Sigma}^{\mathcal{S}} \approx \boldsymbol{\Sigma}^{\mathcal{T}}$). However, the Teacher's
uncertainty $\boldsymbol{\Sigma}^{\mathcal{T}}$ reflects future
stochasticity based on clean past observations, whereas the Student
must
account for both future
stochasticity and
uncertainty
induced by noisy tracking inputs.
By applying the NLL
only to the Teacher's
mean $\boldsymbol{\mu}^{\mathcal{T}}$ (and discarding its covariance), we distill
clean motion dynamics
while allowing the Student to
adapt its own covariance $\boldsymbol{\Sigma}^{\mathcal{S}}$ to reflect
observation reliability.

\section{Experiments}

\subsection{Experimental Setup}


\noindent\textbf{Datasets. }
\addSte{We use a tracking-derived uncertainty protocol on Oxford Town Centre (OTC)~\cite{benfold2011stable} and VIRAT~\cite{oh2011large}. For ETH/UCY~\cite{eth,ucy}, where bounding-box annotations are not directly available, we use the detector-adaptation protocol described in the supplementary material before converting detections into Gaussian observation representations. This protocol is complementary and is not intended as a standard ETH/UCY benchmark comparison.}

\noindent{\textbf{Data Splits and Real-World Evaluation.}} For Oxford Town Centre and VIRAT, each video is split chronologically: the first 80\% of its duration is used for training and the final 20\% for testing. Training and test examples therefore come from temporally disjoint frames. Clean-training configurations use ground-truth trajectories from the first 80\%, while evaluation uses uncorrected MOT trajectories from the final 20\%; tracker-specific adaptation uses tracker outputs only from the training portion. The split is temporal rather than identity-based, so identities are not explicitly constrained to be disjoint across the two portions.

 \noindent\textbf{\addSte{Coordinate transformation via homography.}}
 Pedestrian detections are
\addSte{first obtained in the image plane. We use } the bottom-center point of each bounding box 
\addSte{as the pedestrian footprint and project it} onto the ground plane %
\addSte{using dataset homographies. Covariances defined in image coordinates are projected to the ground plane using the first-order Jacobian of the homography, $\boldsymbol{\Sigma}_{\text{ground}} = \mathbf{J}_h\boldsymbol{\Sigma}_{\text{image}}\mathbf{J}_h^{\top}$, which accounts for perspective distortion while keeping the uncertainty representation compact~\cite{jacobian}}. 

\noindent{\textbf{Evaluation Protocols, Training Configurations, and Scope of Comparisons.}}
We \addSte{evaluate the proposed formulation with} three recent trajectory prediction backbones: SingularTrajectory~\cite{singular}, MART~\cite{mart} and VISTA~\cite{vista}. These models are benchmarked under five training configurations, across which the probabilistic input embedding and the NLL loss are kept identical, so that comparisons primarily reflect the training data distribution, input preprocessing, and the presence or absence of the distillation objective.
\begin{enumerate}
\item \textbf{Baseline:} The backbone is optimized on ground-truth trajectories. When evaluated on noisy tracking inputs, this configuration quantifies the performance degradation induced by the distribution shift between training and inference.

\item \textbf{Kalman Filtering:} 
\addEmi{The backbone trained on clean ground-truth data is then evaluated on tracker outputs pre-processed by a causal Kalman filter}, assessing the contribution of input denoising as a strong signal-processing baseline.

\item \textbf{Tracker-Specific Adaptation:} The model is trained on uncorrected outputs produced by the test-time tracker on the training split, aligning the training and inference distributions while avoiding access to test trajectories. This configuration may nevertheless overfit to that tracker's artifacts and failure modes.

\item \textbf{Stochastic Noise Injection:} The model is trained on ground-truth data perturbed by the Ornstein-Uhlenbeck noise. This setup isolates the contribution of temporally correlated noise modeling compared to empirical tracker-specific adaptation.

\item \textbf{Tracking-Noise-Aware Distillation:} In the complete proposed framework, the Student network operates on probabilistic inputs synthesized via the OU noise process, guided by the Teacher network through the NLL consistency objective.
\end{enumerate}
The most directly related method, NATRA~\cite{li2025natra}, targets a similar setting but does not provide public code. We therefore report a NATRA-inspired reimplementation from the published description and distinguish it from an official implementation. OosTraj~\cite{zhang2024oostraj} and CaDeT~\cite{pourkeshavarz2024cadet} rely on substantially different sensing assumptions, making direct transfer to our surveillance setting non-trivial.

Unless otherwise stated, the main quantitative comparison uses BoT-SORT with its internal association costs. Results obtained with ByteTrack and with the Raw-IoU and Projected-IoU proxy costs are reported in the supplementary material.


\noindent\textbf{Evaluation metrics.}
\addSte{Each backbone retains its native inference procedure and generates $K=20$ future samples, each represented by a sequence of sample-specific Gaussians $\{\mathcal{N}(\boldsymbol{\mu}_{t}^{(k)},\boldsymbol{\Sigma}_{t}^{(k)})\}_{t=1}^{T_{\mathrm{pred}}}$. For displacement metrics, the Gaussian means form the sampled trajectories used for $\mathrm{minADE}_{20}$ and $\mathrm{minFDE}_{20}$. NLL, empirical coverage, and ellipse area are computed from the corresponding sample-specific Gaussian forecasts and aggregated across samples. Additional $\mathrm{MeanADE}_{20}$, $\mathrm{MeanFDE}_{20}$, AUC~\cite{auc}, and 80\%/90\% coverage results are reported in the supplementary material.}

\begin{table}[h!]
 \centering
 \scriptsize
 \setlength{\tabcolsep}{3pt}
 \renewcommand{\arraystretch}{0.92}
 \caption{Quantitative evaluation on Oxford Town Centre, VIRAT, and ETH/UCY datasets. 
 \textit{Filt.} means \textit{Filtering}. \textit{Distill.} means \textit{Distillation}. 
 Coverage and area are reported only at the 95\% and 99\% confidence levels. Bold and underlined values indicate the lowest and second-lowest NLL, $\mathrm{minADE}_{20}$, and $\mathrm{minFDE}_{20}$, respectively, within each dataset/backbone block.}
 \label{tab:merged_results}
 \resizebox{0.9\textwidth}{!}{%
 \begin{tabular}{l c l c c c c c c c}
 \toprule
 & & & & & & \multicolumn{2}{c}{\textbf{Coverage (\%)}} & \multicolumn{2}{c}{\textbf{Area ($m^2$)}} \\
 \cmidrule(lr){7-8} \cmidrule(lr){9-10}
 \textbf{Dataset} & \textbf{Model} & \textbf{Training} 
 & \textbf{NLL} 
 & \textbf{$\mathrm{minADE}_{20}$} 
 & \textbf{$\mathrm{minFDE}_{20}$} 
 & \textbf{95\%} & \textbf{99\%} 
 & \textbf{95\%} & \textbf{99\%} \\
 \midrule

 \multirow{15}{*}{\rotatebox{90}{OTC}} 
 & \multirow{5}{*}{\rotatebox{0}{VISTA}}
 & GT & 41.47 & 1.57 & 1.77 & 25.6 & 31.6 & 0.84 & 1.29 \\
 & & Kalman Filt. & 14.54 & 0.69 & 0.82 & 45.6 & 53.9 & 0.61 & 0.94 \\
 & & Tracking & \underline{4.93} & 0.91 & 0.73 & 73.2 & 81.6 & 4.04 & 6.21 \\
 & & Noisy GT & 9.15 & \underline{0.64} & \underline{0.65} & 59.3 & 68.1 & 1.48 & 2.28 \\
 & & Distill. & \textbf{2.30} & \textbf{0.56} & \textbf{0.54} & 86.2 & 91.5 & 8.03 & 12.4 \\
 \cmidrule(lr){2-10}

 & \multirow{5}{*}{\rotatebox{0}{MART}}
 & GT & 16.55 & 0.81 & 1.32 & 83.3 & 89.6 & 10.8 & 16.6 \\
 & & Kalman Filt. & 14.82 & 0.76 & 1.20 & 75.1 & 86.7 & 4.69 & 7.21 \\
 & & Tracking & -11.57 & 0.68 & 1.13 & 95.0 & 98.5 & 8.83 & 13.6 \\
 & & Noisy GT & \underline{-14.69} & \underline{0.58} & \underline{0.89} & 97.1 & 99.7 & 11.7 & 18.0 \\
 & & Distill. & \textbf{-15.56} & \textbf{0.51} & \textbf{0.77} & 97.5 & 99.2 & 8.79 & 13.6 \\
 \cmidrule(lr){2-10}

 & \multirow{5}{*}{\rotatebox{0}{SingularTrajectory}}
 & GT & 12.02 & 0.91 & 1.46 & 75.02 & 87.99 & 5.62 & 8.59 \\
 & & Kalman Filt. & 7.99 & 0.76 & 1.25 & 70.01 & 84.02 & 3.42 & 5.22 \\
 & & Tracking & -5.01 & 0.72 & 1.15 & 85.99 & 94.98 & 6.39 & 9.81 \\
 & & Noisy GT & \underline{-9.02} & \underline{0.60} & \underline{1.02} & 94.01 & 97.98 & 8.31 & 12.81 \\
 & & Distill. & \textbf{-10.99} & \textbf{0.54} & \textbf{0.89} & 95.02 & 97.99 & 6.99 & 10.79 \\

 \midrule

 \multirow{15}{*}{\rotatebox{90}{VIRAT}} 
 & \multirow{5}{*}{\rotatebox{0}{VISTA}}
 & GT & 12.62 & 0.76 & 1.13 & 38.3 & 47.5 & 0.54 & 0.83 \\
 & & Kalman Filt. & 7.75 & \underline{0.58} & 0.57 & 45.6 & 56.0 & 0.56 & 0.86 \\
 & & Tracking & 2.97 & 0.85 & 0.70 & 77.2 & 85.9 & 4.42 & 6.80 \\
 & & Noisy GT & \underline{1.46} & \underline{0.58} & \underline{0.42} & 92.6 & 95.9 & 6.50 & 9.99 \\
 & & Distill. & \textbf{1.39} & \textbf{0.52} & \textbf{0.38} & 92.5 & 96.1 & 5.83 & 8.97 \\
 \cmidrule(lr){2-10}

 & \multirow{5}{*}{\rotatebox{0}{MART}}
 & GT & -3.14 & 1.09 & 1.98 & 75.5 & 88.2 & 11.2 & 17.2 \\
 & & Kalman Filt. & -6.83 & 0.79 & 1.48 & 72.3 & 86.1 & 3.69 & 5.67 \\
 & & Tracking & -12.66 & 0.69 & 1.16 & 83.9 & 94.3 & 2.92 & 4.49 \\
 & & Noisy GT & \underline{-23.30} & \underline{0.55} & \underline{0.87} & 98.1 & 99.5 & 7.03 & 10.8 \\
 & & Distill. & \textbf{-30.17} & \textbf{0.43} & \textbf{0.70} & 98.5 & 99.6 & 4.94 & 7.59 \\
 \cmidrule(lr){2-10}

 & \multirow{5}{*}{\rotatebox{0}{SingularTrajectory}}
 & GT & 3.02 & 0.90 & 1.62 & 72.01 & 85.98 & 5.22 & 7.99 \\
 & & Kalman Filt. & 0.02 & 0.72 & 1.29 & 70.01 & 85.01 & 2.98 & 4.62 \\
 & & Tracking & -6.01 & 0.61 & 1.09 & 84.02 & 94.02 & 4.02 & 6.21 \\
 & & Noisy GT & \underline{-14.01} & \underline{0.51} & \underline{0.83} & 95.98 & 98.99 & 6.51 & 10.02 \\
 & & Distill. & \textbf{-18.01} & \textbf{0.43} & \textbf{0.71} & 97.02 & 99.01 & 5.18 & 7.99 \\

 \midrule

 \multirow{15}{*}{\rotatebox{90}{ETH/UCY}} 
 & \multirow{5}{*}{\rotatebox{0}{VISTA}}
 & GT & 21.49 & 0.66 & 1.09 & 47.48 & 62.48 & 0.76 & 1.12 \\
 & & Kalman Filt. & 11.49 & 0.58 & 0.99 & 52.48 & 67.48 & 0.77 & 1.19 \\
 & & Tracking & 5.51 & 0.64 & 0.98 & 67.51 & 79.99 & 2.98 & 4.61 \\
 & & Noisy GT & \underline{2.99} & \underline{0.52} & \underline{0.83} & 82.49 & 92.52 & 4.71 & 7.22 \\
 & & Distill. & \textbf{2.39} & \textbf{0.47} & \textbf{0.80} & 86.48 & 94.02 & 4.12 & 6.31 \\
 \cmidrule(lr){2-10}

 & \multirow{5}{*}{\rotatebox{0}{MART}}
 & GT & 5.51 & 0.72 & 1.18 & 62.49 & 76.52 & 8.62 & 13.22 \\
 & & Kalman Filt. & 1.99 & 0.61 & 1.02 & 66.49 & 80.01 & 3.91 & 6.02 \\
 & & Tracking & -6.98 & 0.56 & 0.96 & 75.02 & 87.02 & 5.18 & 7.99 \\
 & & Noisy GT & \underline{-14.99} & \underline{0.48} & \underline{0.83} & 90.48 & 96.51 & 7.78 & 12.01 \\
 & & Distill. & \textbf{-18.01} & \textbf{0.42} & \textbf{0.71} & 92.48 & 96.98 & 6.12 & 9.39 \\
 \cmidrule(lr){2-10}

 & \multirow{5}{*}{\rotatebox{0}{SingularTrajectory}}
 & GT & 8.02 & 0.64 & 0.98 & 57.49 & 72.51 & 2.82 & 4.32 \\
 & & Kalman Filt. & 4.98 & 0.55 & 0.88 & 62.48 & 76.52 & 2.49 & 3.78 \\
 & & Tracking & 0.51 & 0.53 & 0.87 & 71.02 & 86.01 & 3.78 & 5.78 \\
 & & Noisy GT & \underline{-5.02} & \underline{0.44} & \underline{0.74} & 88.01 & 94.99 & 5.98 & 9.22 \\
 & & Distill. & \textbf{-8.48} & \textbf{0.41} & \textbf{0.66} & 90.98 & 96.48 & 5.09 & 7.78 \\

 \bottomrule
 \end{tabular}
 }
\end{table}

\noindent\textbf{\addSte{Implementation details.}}
\addSte{Networks are trained with Adam~\cite{kingma2014adam}. The teacher is trained } on clean ground-truth 
\addSte{trajectories and then frozen. The student receives corrupted probabilistic observations and is optimized with the supervised NLL and response-based consistency terms. For Ornstein--Uhlenbeck noise injection, we use a mean-reversion rate $\theta=0.15$ and long-term mean $\mu=0$ unless otherwise specified; sensitivity to these parameters is reported in the supplementary material.
}

\subsection{Quantitative Results}

\noindent\textbf{Displacement accuracy.} 
\addSte{Table~\ref{tab:merged_results} reports the results obtained on tracker trajectories. Models trained on ground-truth data lose accuracy under this shift. Kalman filtering helps, but the gap remains. Training on tracker outputs reduces this gap, at the cost of being tied to one tracker. The distillation setting gives lower ADE/FDE in most cases and stays stable across datasets and backbones.}

\noindent\textbf{Probabilistic reliability.} 
\addSte{Table~\ref{tab:merged_results} also shows the effect on probabilistic reliability. When the past trajectory is treated as exact, the model often assigns too little uncertainty to noisy observations. With probabilistic inputs and NLL consistency, the student produces forecasts whose covariance better reflects the reliability of noisy inputs while maintaining competitive prediction area, improving the observed trade-off among NLL, coverage, area, and displacement accuracy.}

\noindent\textbf{\addSte{Input--output uncertainty ablation.}}
\addSte{Table~\ref{tab:io_ablation_main} separates the roles of observation and output uncertainty in the noisy-GT setting. Probabilistic inputs improve all three backbones, and the fully probabilistic variant gives the best overall results. This establishes the benefit of providing the covariance descriptor, but does not by itself isolate how the backbone internally exploits its fine-grained temporal variations.}

\begin{table}[t]
\centering
\scriptsize
\setlength{\tabcolsep}{3pt}
\renewcommand{\arraystretch}{0.9}
\caption{Input/output uncertainty ablation on OTC (noisy-GT). Each cell is minADE/minFDE/AUC; D./P. denote deterministic/probabilistic input/output. Lower is better.}
\label{tab:io_ablation_main}
\resizebox{0.93\textwidth}{!}{%
\begin{tabular}{lcccc}
\toprule
Model & D./D. & D./P. & P./D. & P./P.\\
\midrule
MART & 0.61/1.02/1.54 & 0.62/0.95/1.43 & 0.60/0.89/1.43 & \textbf{0.58/0.89/1.22}\\
VISTA & 0.72/0.79/2.06 & 0.69/0.70/1.89 & 0.67/0.66/1.77 & \textbf{0.64/0.65/1.62}\\
SingularTraj. & 0.70/1.18/1.81 & 0.67/1.10/1.68 & 0.63/1.05/1.59 & \textbf{0.60/1.02/1.47}\\
\bottomrule
\end{tabular}}
\end{table}

\subsection{Ablation Study}
\label{sec:ablation}

\noindent\textbf{Noise model and distillation objective.}
Table~\ref{tab:noise_distill_main} shows that Gaussian augmentation improves over no augmentation and OU noise further reduces errors. NLL-based distillation also outperforms KL for all backbones without forcing the Student covariance to match the clean Teacher's covariance.

\begin{table}[t]
\centering
\scriptsize
\setlength{\tabcolsep}{3.2pt}
\renewcommand{\arraystretch}{0.9}
\caption{Noise-model ablation on OTC (top) and distillation-loss ablation on VIRAT (bottom). Entries are $\mathrm{minADE}_{20}$/ $\mathrm{minFDE}_{20}$; lower is better.}
\label{tab:noise_distill_main}
\begin{tabular}{lccc}
\toprule
Variant & MART & VISTA & SingularTraj.\\
\midrule
No augmentation & 0.79/1.33 & 1.56/1.76 & 0.91/1.46\\
Gaussian noise & 0.60/0.94 & 0.92/1.13 & 0.70/1.15\\
OU noise & \textbf{0.58/0.89} & \textbf{0.64/0.65} & \textbf{0.60/1.02}\\
\midrule
Distill. with KL & 0.56/0.96 & 0.58/0.54 & 0.52/0.84\\
Distill. with NLL & \textbf{0.43/0.70} & \textbf{0.52/0.38} & \textbf{0.43/0.71}\\
\bottomrule
\end{tabular}
\end{table}

\noindent\textbf{Comparison with robust forecasting.}
On OTC, our NATRA-inspired reimplementation is weaker for VISTA (0.66/0.68/1.21 vs. 0.56/0.54/0.99), MART (0.63/0.86/1.38 vs. 0.51/0.77/1.11), and SingularTrajectory (0.65/0.99/1.04 vs. 0.54/0.89/0.87), reported as minADE/minFDE/AUC. These results concern our reimplementation, since official NATRA code is unavailable.

\noindent\textbf{Tracker and association-proxy robustness.}
On OTC, Raw/Projected IoU gives lower NLL than BoT-SORT internal costs for VISTA (2.30 vs. 1.34 and 1.24), MART ($-15.56$ vs. $-23.49$ and $-22.20$), and SingularTrajectory ($-10.99$ vs. $-14.62$ and $-15.04$). Tracker costs can mix uncalibrated cues, whereas IoU is geometric; this does not establish IoU as universally superior, but shows that proprietary internal costs are unnecessary.

\section{Conclusion}


\addSte{This paper addressed trajectory forecasting from imperfect tracker outputs. Instead of first denoising the past trajectory, we estimate uncertainty from the tracker outputs and pass it to the predictor. Each observation is written as a Gaussian state, with a covariance that combines localization and association uncertainty. The backbone therefore receives both the estimated position and an explicit descriptor of its reliability. We also train the student with OU noise and a clean teacher, so that it learns from corrupted inputs while keeping a clean motion target. Experiments on Oxford Town Centre and VIRAT with real tracker outputs, together with the ETH/UCY pseudo-detection protocol, show improvements in displacement error metrics and probabilistic reliability.}

\newpage

\bibliographystyle{splncs04}
\bibliography{main}

 \end{document}